\documentclass{article}

\usepackage{graphicx}
\usepackage{subcaption}
\usepackage{wrapfig}

\usepackage[most]{tcolorbox}
\usepackage[T1]{fontenc}
\usepackage{multirow}

\usepackage[utf8]{inputenc} 
\usepackage[T1]{fontenc}    
\usepackage{hyperref}       
\usepackage{url}            
\usepackage{booktabs}       
\usepackage{amsfonts}       
\usepackage{nicefrac}       
\usepackage{microtype}      
\usepackage{xcolor}         
\usepackage{amsmath}
\usepackage[capitalize]{cleveref}
\usepackage{amsmath}
\hypersetup{
    colorlinks,
    linkcolor={red!50!black},
    citecolor={blue!50!black},
    urlcolor={blue!80!black}
}
\usepackage[utf8]{inputenc} 
\usepackage[T1]{fontenc}    
\usepackage{hyperref}       
\usepackage{url}            
\usepackage{booktabs}       
\usepackage{amsfonts}       
\usepackage{nicefrac}       
\usepackage{microtype}      
\usepackage{xcolor}         
\usepackage{graphicx}
\usepackage{amsmath,amssymb}
\usepackage{hyperref}       
\usepackage[capitalize]{cleveref}      
\usepackage{url}            
\usepackage{booktabs}       
\usepackage{amsfonts}       
\usepackage{nicefrac}       
\usepackage{microtype}      
\usepackage{xcolor}         
\usepackage{float}
\usepackage{multirow}
\usepackage{booktabs}
\usepackage{graphicx}

\usepackage{textgreek}
\usepackage{tikz} 
\usepackage{pgfplots}

\usepackage{adjustbox}            

\usepackage[capitalize]{cleveref}
\usepackage{hyperref} 
\usepackage{url}            
\usepackage{booktabs}       
\usepackage{amsfonts}    
\usepackage{times}
\usepackage{epsfig}
\usepackage{graphicx}
\usepackage{amsmath}
\usepackage{amssymb}
 \usepackage[preprint]{neurips_2026}

\title{Test-Time Training for Modality Order Consistency in Vision-Language Models}

\author{%
  Aditi Gupta \\
  University of Chicago\\
  \And
  Yossi Gandelsman \\
  Reve \\
}

\begin{document}

\maketitle

\begin{abstract}
We find that vision-language models are sensitive to a specific semantically irrelevant change: the order in which the image and question are presented. Across three models and three benchmarks, image-first prompting consistently outperforms question-first prompting, revealing a repeatable modality order failure. We use this gap to design an order-consistent test-time training method. Our method substantially closes the modality-order gap across all evaluated settings. Surprisingly, it also yields consistent improvements in the stronger image-first branch over the baseline, hence bootstrapping both orderings toward mutual consistency. Activation patching localizes the ordering failure to a narrow mid-network region where representations diverge sharply between prompt orders. We find that the test-time training method repairs this misalignment across layers. Together, our results identify modality-order sensitivity as a circuit-level failure in VLMs and demonstrate that simple, asymmetric test-time adaptation can effectively mitigate it and even improve performance over the baseline. Code is available at \href{https://aditiii12.github.io/ttt-for-vlms/}{link}.
\end{abstract}

\section{Introduction}
Vision-language models (VLMs) have advanced rapidly in recent years, achieving strong performance across a broad range of multimodal tasks, including visual question answering, image captioning, and multimodal reasoning~\citep{bai2025qwen3, chen2024internvl, alayrac2022flamingo, li2022blip, liu2023visual}. These models process images and text jointly, but the precise mechanisms by which visual and linguistic information are integrated remain poorly understood. In particular, it is unclear how the relative ordering of modalities in the input affects internal computation, and whether models remain robust when the semantic content and underlying multimodal task remain unchanged.

We identify that modern VLMs are very sensitive to modality order in prompts: presenting the same image and question in different orders, that is, swapping the textual query and the image component of the prompt, produces substantially different predictions on a significant fraction of examples across multiple model families and datasets. We find that image-first prompting consistently outperforms query-first with accuracy gaps ranging from 6 to 26 percentage points. A semantically irrelevant perturbation thus induces a repeatable and structured performance asymmetry, suggesting a systematic difference in the internal computations elicited by two prompt orderings.

We propose a simple test-time training (TTT) procedure to repair this inconsistency. For each test instance, we construct both orderings and treat the stronger image-first variant as a teacher signal. We then perform a small number of gradient steps that align the question-first prediction distribution toward the image-first one, using a KL divergence test-time objective that keeps the teacher fixed. Our procedure instead uses the stronger computation as an anchor and selectively repairs the weaker one, without requiring labeled data, external supervision, or task-specific fine-tuning at test time. 

After a small number of test-time gradient steps, our method substantially closes the question-first accuracy gap across all three models and datasets. More surprisingly, image-first accuracy also improves slightly in most settings, suggesting that the adaptation procedure does not merely overfit to one ordering but produces a more internally consistent model. These results hold across model families and benchmark types, indicating that the phenomenon and its repair are not artifacts of any particular architecture, dataset, or evaluation setting.

To understand where the ordering failure arises and how repair works internally, we combine activation patching with a TTT ablation over contiguous layer windows, where each window restricts gradient updates to a fixed block of transformer layers. Activation patching reveals that the failure is causally concentrated in a narrow mid-network band: replacing query-first hidden states with image-first states in early layers has almost no effect, while patching within a specific mid-layer window sharply restores accuracy. Our window ablation of test-time training is consistent with this
localization: adaptation targeted at mid-network layers achieves faster convergence and higher question-first accuracy than adaptation at earlier or later layers. We additionally find that the asymmetric objective is essential: symmetric alternatives that treat both orderings as equally
reliable recover substantially less of the ordering gap, confirming that the directionality must be reflected in the adaptation objective.

Swapping the order of image and question should not materially change a vision-language model's answer, yet in practice it often does. We show that this semantically irrelevant prompt perturbation induces a repeatable circuit-level failure in VLMs, that the failure is causally localizable, and that a directional test-time repair can use the stronger image-first branch to effectively repair predictions produced by the weaker question-first branch.

In summary, our key contributions are :
\begin{itemize}
    \item We identify a circuit-level failure mode in vision-language models induced by modality ordering: identical image-question pairs can produce substantially different predictions when the order of the two modalities is swapped.
    \item Using activation patching, we show that the ordering failure is not diffuse, but is causally localized to a narrow depth regime.
    \item We propose a mechanism-guided asymmetric test-time repair procedure that uses the stronger image-first variant as teacher to repair the weaker question-first variant, closing most of the ordering gap while largely preserving the stronger image-first behavior.
\end{itemize}

\section{Related Work}
\label{related_work}

\paragraph{Vision Language Models.}
Modern vision-language models integrate visual and linguistic information through a shared transformer backbone, either via cross-attention mechanisms \citep{alayrac2022flamingo, li2023blip} or by projecting visual features into the language model's token space \citep{liu2023visual, dai2023instructblip}. Despite strong performance across multimodal benchmarks, the precise mechanisms by which these models fuse modalities remain poorly understood. Prior work has identified sensitivity to prompt formatting~\citep{huang2023language}, and recent studies suggest that VLMs may rely disproportionately on text over visual input in certain settings~\citep{deng2025words} and that a performance gap persists between analogous visual and textual tasks across model families~\citep{samedifftask}. Our work contributes to this understanding by identifying modality order as a structured and repeatable source of inconsistency, and we propose a
test-time training method that closes this gap without harming stronger ordering, and surprisingly, slightly improves it.

\paragraph{Test-time training.}
Our method builds on test-time training~\citep{sun2020test,gandelsman2022testtime}, which
adapts model parameters at inference using self-supervised signals from
each test instance without labeled data. Subsequent methods minimize entropy~\citep{wang2020tent}, enforce augmentation consistency~\citep{zhang2022memo}, or adapt vision-language models via prompt learning~\citep{zhou2022learning, zhou2022conditional, khattak2023maple}.  Most of these settings focus on the adapted distribution. Our method produces a qualitatively different effect: because the teacher is recomputed from updated weights at each step, improvements in question-first feed back into a sharper image-first teacher, bootstrapping both orderings toward mutual consistency. \citet{chou2026test} propose a symmetric consistency objective for VLMs. Through empirical results, we show that treating both orderings as equally reliable recovers substantially less of the observed modality-order performance gap.
More broadly, consistency constraints have been used to preserve underlying
content across transformations
\citep{zhu2017unpaired}. Our setting similarly treats the two modality
orderings as semantics-preserving views, but uses a directional consistency
objective because image-first prompting is empirically more reliable.
\paragraph{Prompt Sensitivity.}
LLMs are sensitive to prompt formatting, including option order~\citep{zhao2021calibrate}, instruction phrasing~\citep{mizrahi2024state}, and the ordering of in-context examples~\citep{lu2022fantastically}. In VLMs, sensitivity to prompt format has been documented across visual question answering~\citep{liu2024mmbench} and multimodal reasoning~\citep{chen2024we} tasks. Closest to our work,
\citet{deng2025words} and \citet{hejabi2025flip} document that image-first prompting outperforms question-first across models and tasks.
These works establish the phenomenon. Our work goes further by localizing the failure via activation patching, and proposing a principled test-time repair that can even improve the stronger ordering.

\section{Test-Time Training for Order Consistency}
\label{sec:ttt_order_consistency}

In this section, we introduce a label-free test-time training method for reducing modality-order sensitivity in VLMs. We first document the modality-ordering failure, then present our asymmetric adaptation objective and evaluate its effect on both prompt orders.

\subsection{Modality-Ordering Failure in Vision-Language Models}
\label{sec:ordering_failure}

\begin{figure}[t]
    \centering
    \includegraphics[width=\linewidth]{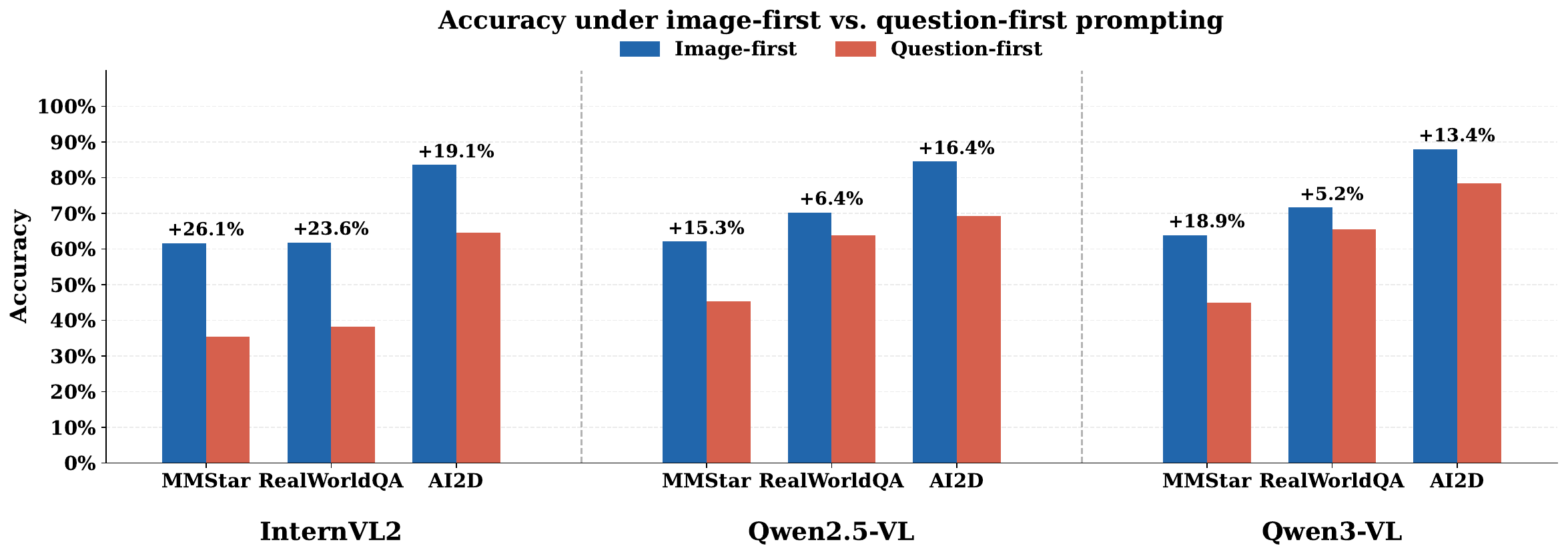}
    \caption{\textbf{Accuracy under image-first and question-first prompting.}
    Image-first prompting consistently outperforms question-first prompting across datasets and model families.}
    \label{fig:ordering_gap}
\end{figure}

For each benchmark instance, we evaluate the same image-question pair under two prompt orders: image-first and question-first. \Cref{fig:ordering_gap} shows that image-first prompting achieves higher accuracy across the evaluated datasets and model families. The two orderings also disagree on a substantial fraction of examples, indicating that modality order changes model behavior even when the underlying image, question, answer choices, and semantic task remain unchanged.

The gap is largest on harder multimodal reasoning benchmarks, but remains visible on natural visual question-answering tasks. Since the two prompts contain the same image and question, the accuracy gap must arise from differences in the computation induced by the two input orders.

\subsection{Test-Time Training Consistency Objective}
\label{sec:asymmetric_ttt}

The asymmetry in \cref{fig:ordering_gap} motivates a directional adaptation strategy. Rather than forcing both prompt orders toward a common midpoint, we use the stronger image-first prediction as a detached teacher and adapt the question-first prediction toward it. For each test instance, we perform a small number of gradient updates using an asymmetric consistency loss between the two output distributions. This makes the update label-free, instance-specific, and targeted at the weaker ordering identified above.

\paragraph{Setup.}
Let $f_{\theta}$ denote a vision-language model with parameters $\theta$. For a test instance
\[
s = (x, q, \mathcal{A}),
\]
where $x$ is an image, $q$ is a question, and $\mathcal{A}$ is the set of answer candidates, we define two prompt orderings:
\[
o_{\mathrm{IF}} = \text{image-first},
\qquad
o_{\mathrm{QF}} = \text{question-first}.
\]
The model produces logits over answer candidates:
\[
z_{\mathrm{IF}}(\theta; s) = f_{\theta}(s, o_{\mathrm{IF}}),
\qquad
z_{\mathrm{QF}}(\theta; s) = f_{\theta}(s, o_{\mathrm{QF}}).
\]
We convert logits to predictive distributions,
\[
p_{\mathrm{IF}}(\theta; s) = \operatorname{softmax}(z_{\mathrm{IF}}(\theta; s)),
\qquad
p_{\mathrm{QF}}(\theta; s) = \operatorname{softmax}(z_{\mathrm{QF}}(\theta; s)).
\]

For each test instance, we initialize from the pretrained parameters $\theta^{(0)}=\theta_0$ and perform $K$ adaptation steps, where each adaptation step is one gradient update on that same instance. At step $k$, we recompute the image-first distribution using the current parameters and detach it:
\[
\bar{p}_{\mathrm{IF}}^{(k)}(s)
=
\operatorname{stopgrad}\!\left(
p_{\mathrm{IF}}(\theta^{(k)}; s)
\right).
\]
Thus, the teacher is fixed within each gradient step but updated across steps. We use this moving-teacher variant by default; \cref{tab:frozen_vs_moving} shows that recomputing the teacher substantially outperforms keeping the initial image-first prediction fixed. After evaluating the adapted model on both prompt orders, we reset the parameters to $\theta_0$ before processing the next example. This reset ensures that the adaptation remains label-free, instance-specific, and independent of all previous examples.

\paragraph{Objective.}

At step $k$, our alignment loss is
\[
\mathcal{L}_{\mathrm{align}}(\theta^{(k)}; s)
=
D_{\mathrm{KL}}
\left(
\bar{p}_{\mathrm{IF}}^{(k)}(s)
\;\middle\|\;
p_{\mathrm{QF}}(\theta^{(k)}; s)
\right).
\]
We perform gradient updates of the form
\[
\theta^{(k+1)}
=
\theta^{(k)}
-
\eta \nabla_{\theta^{(k)}}
\mathcal{L}_{\mathrm{align}}(\theta^{(k)}; s),
\qquad
k=0,\ldots,K-1.
\]
After adaptation, we evaluate the updated model on the target prompt order and then reset to $\theta_0$ before beginning adaptation on the subsequent test instance.

This objective is asymmetric: gradients flow through the question-first prediction, while the image-first distribution acts as a detached teacher at each step. This encodes the empirical direction of the failure in \cref{fig:ordering_gap}; image-first provides the anchor, and question-first is adapted toward it.

In principle, the update can be applied to all model parameters. In practice, our mechanistic analysis in \cref{sec:analysis} shows that the modality-ordering failure is concentrated in a specific depth range. We therefore compare adaptation over early, middle, and late layer windows, and find that mid-layer windows give the best trade-off between question-first recovery and image-first performance. This turns the causal analysis into a mechanism-guided test-time repair procedure, rather than a generic model-wide adaptation step applied uniformly throughout the entire network.

\section{Experiments}
\label{sec:experiments}
We evaluate whether asymmetric order-consistency TTT reduces the question-first accuracy gap identified in \cref{sec:ordering_failure}. Across three VLMs spanning two model families and three benchmarks, our method consistently improves question-first accuracy while slightly improving performance under the stronger image-first ordering, without introducing a trade-off between the two.


\begin{table}[t]
\centering
\small
\resizebox{\textwidth}{!}{%
\begin{tabular}{llcccccc}
\toprule
\textbf{Model} & \textbf{Dataset}
& \textbf{\shortstack{Image-first\\baseline}}
& \textbf{\shortstack{Question-first\\baseline}}
& \textbf{\shortstack{Question-first\\TTT}}
& \textbf{$\Delta_{\text{QF}}$}
& \textbf{\shortstack{Image-first\\TTT}}
& \textbf{$\Delta_{\text{IF}}$} \\
\midrule

\multirow{3}{*}{\textbf{InternVL2-8B}}
 & \textbf{MMStar}      & 61.6 & 35.5 & 61.5 & +26.1 & 62.3 & \textbf{+0.7} \\
 & \textbf{RealWorldQA} & 61.8 & 38.2 & 60.0 & +21.8 & 62.1 & \textbf{+0.3} \\
 & \textbf{AI2D}        & 83.6 & 64.5 & 83.0 & +18.5 & 83.6 & +0.0 \\
\midrule

\multirow{3}{*}{\textbf{Qwen2.5-VL-7B}}
 & \textbf{MMStar}      & 62.1 & 45.3 & 58.7 & +13.4 & 62.7 & \textbf{+0.5} \\
 & \textbf{RealWorldQA} & 67.8 & 60.0 & 64.1 & +4.1  & 68.4 & \textbf{+0.5} \\
 & \textbf{AI2D}        & 84.5 & 69.3 & 82.3 & +12.9 & 84.5 & +0.0 \\
\midrule

\multirow{3}{*}{\textbf{Qwen3-VL-8B}}
 & \textbf{MMStar}      & 63.9 & 44.9 & 64.4 & +19.5 & 65.1 & \textbf{+1.1} \\
 & \textbf{RealWorldQA} & 71.6 & 65.5 & 70.7 & +5.2  & 72.8 & \textbf{+1.2} \\
 & \textbf{AI2D}$^\dagger$
                           & 88.0 & 78.5 & 87.0 & +8.5 & 88.5 & \textbf{+0.5} \\
\bottomrule
\end{tabular}%
}

\vspace{1mm}
\caption{\textbf{Main results.}
Asymmetric order-consistency TTT improves question-first (QF) accuracy across all evaluated model--dataset pairs while maintaining or slightly improving image-first (IF) accuracy. Results are reported at a fixed model-specific step count: $K{=}5$ for InternVL2-8B, $K{=}4$ for Qwen2.5-VL-7B, and $K{=}3$ for Qwen3-VL-8B.}
\label{tab:main_results}
\end{table}

\begin{figure*}[t]
    \centering

    \begin{subfigure}[t]{0.475\textwidth}
        \centering
        \includegraphics[
            width=\linewidth,
            height=4.9cm
        ]{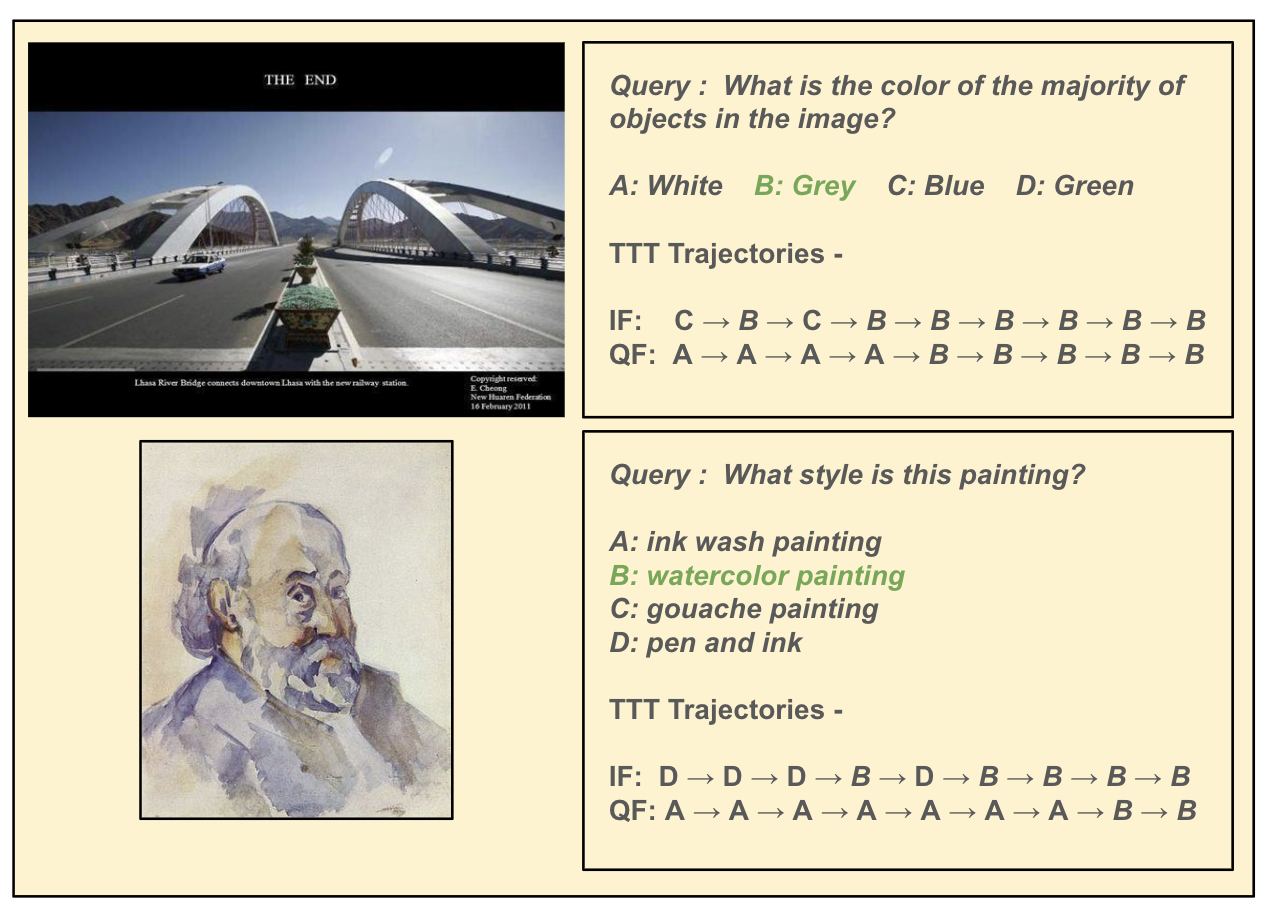}
        \label{fig:qual_mmstar}
    \end{subfigure}
    \hspace{0.005\textwidth}
    \begin{subfigure}[t]{0.475\textwidth}
    \centering
    \raisebox{0.07cm}{%
        \includegraphics[
            width=\linewidth,
            height=4.83cm
        ]{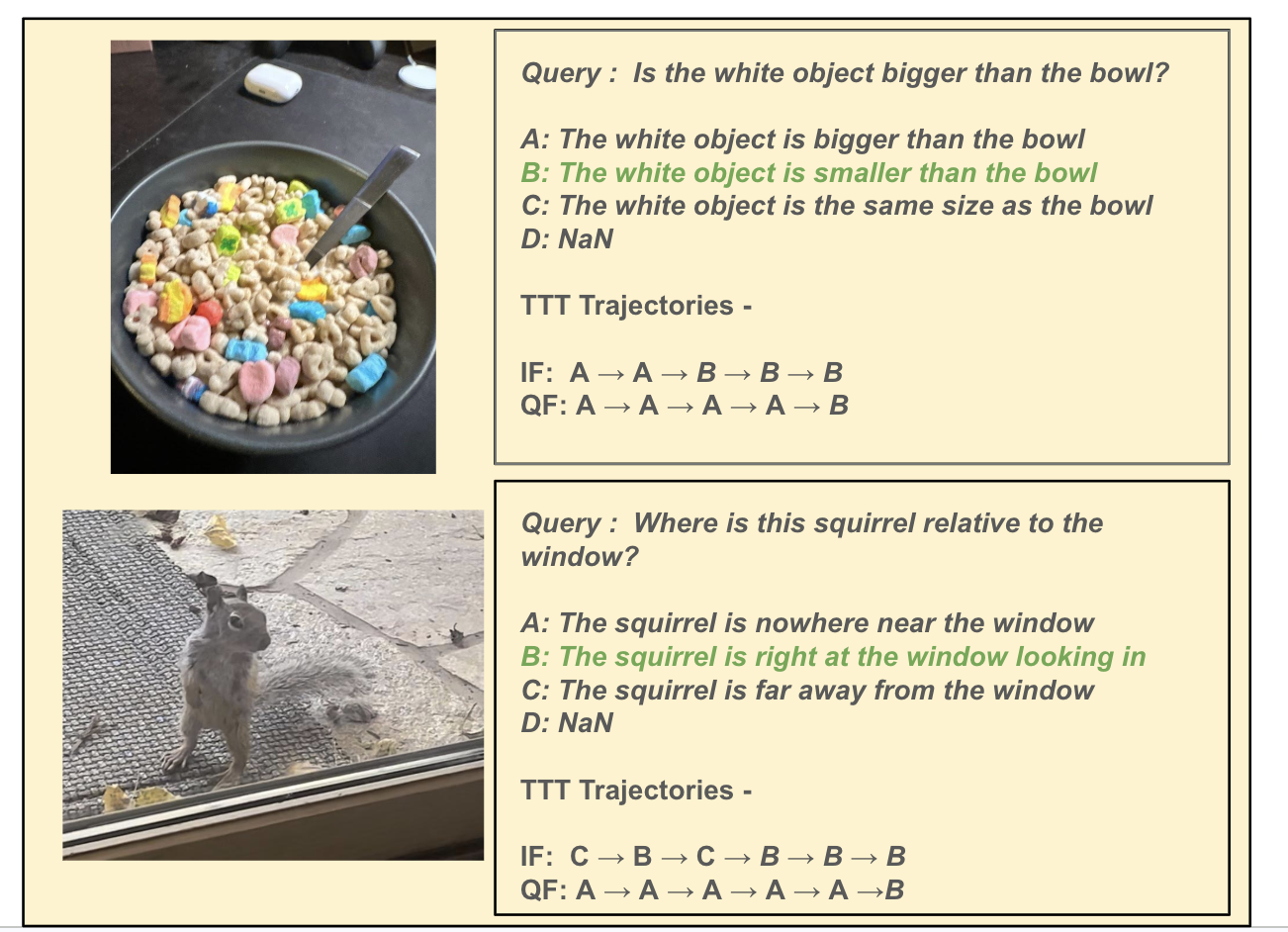}%
    }
    \label{fig:qual_rw}
\end{subfigure}

    \caption{\textbf{Qualitative TTT trajectories.}
    Each panel shows image-first (IF) and question-first (QF) predictions across TTT steps; the correct answer is highlighted in green. In these examples, IF reaches the correct answer earlier, while QF catches up over the adaptation trajectory.}
    \label{fig:qual_examples}
\end{figure*}
\subsection{Experimental Setup}
\label{sec:experimental_setup}

\paragraph{Models.}
We evaluate on three recent open-source vision-language models spanning two model families. \textbf{InternVL2-8B}~\citep{chen2024internvl} uses a
ViT-based visual encoder projected into the language model's token space
via an MLP connector. \textbf{Qwen2.5-VL-7B}~\citep{qwen2.5} and
\textbf{Qwen3-VL-8B}~\citep{bai2025qwen3} use a native-resolution vision encoder with dynamic image tiling. All models are instruction-tuned and evaluated in zero-shot mode without task-specific fine-tuning.

\paragraph{Datasets.}
We evaluate on MMStar ~\citep{chen2024we}, RealWorldQA, and AI2D ~\citep{kembhavi2016diagram}. These benchmarks vary in visual dependency and language-prior strength. MMStar contains 1,500 samples filtered to require visual understanding. RealWorldQA contains 765 real-world perception questions. AI2D contains 3,088 diagram questions and has stronger language priors; for example, Qwen2.5-VL obtains $59.3\%$ text-only accuracy on AI2D compared to $26.6\%$ on MMStar. This provides a useful control: the ordering gap is largest on visually demanding benchmarks, but remains present even when language priors are stronger (Appendix~\ref{app:textonly}).

\paragraph{Prompt orders.}
For each instance, we evaluate two prompts that contain the same image, question, and answer candidates. The image-first prompt places the image before the question; the question-first prompt places the question and options before the image. Both prompts use the model's standard chat template, and only the relative position of the image token block is changed.

\paragraph{Test-time adaptation.}
For each test instance, we follow the protocol described in ~\cref{sec:asymmetric_ttt}. Adaptation targets only the language model backbone parameters within the selected layer window; vision encoder parameters are held fixed throughout. We use SGD with momentum $0.9$. We set the learning rate to $\eta=5\times 10^{-4}$ for InternVL2-8B and $\eta=5\times 10^{-5}$ for Qwen2.5-VL and Qwen3-VL. For the main results in \cref{tab:main_results}, we report performance at a fixed model-specific step count selected on a held-out calibration split: $K=5$ for InternVL2-8B, $K=4$ for Qwen2.5-VL-7B, and $K=3$ for Qwen3-VL-8B. No labels from the test instance are used during adaptation.
\subsection{Main Results}
\label{sec:main_results}

\paragraph{TTT closes the question-first gap.}
\Cref{tab:main_results} shows that asymmetric order-consistency TTT improves question-first accuracy in all nine model--dataset settings. The gains range from $+5.2$ to $+26.1$ percentage points. In several settings, the adapted question-first accuracy nearly reaches the image-first baseline: InternVL2 improves from $35.5\%$ to $61.5\%$ on MMStar, compared to an image-first baseline of $61.6\%$, and from $38.2\%$ to $60.0\%$ on RealWorldQA, exceeding the image-first baseline of $61.8\%$. These results show that a large fraction of the modality-ordering gap can be repaired at test time without labels.

\paragraph{Image-first accuracy is preserved or improves.}
Although the adaptation objective targets the weaker question-first ordering, image-first accuracy does not decrease in any setting. It improves in seven out of nine model--dataset pairs and remains unchanged in the other two. The gains are small but consistent, ranging from $+0.0$ to $+1.2$ percentage points, with the largest gains on Qwen3-VL RealWorldQA and MMStar. This indicates that aligning question-first predictions to the image-first teacher does not trade off against the stronger ordering and can slightly improve both prompt orders.

\subsection{Ablations}
\label{sec:ablations}

\paragraph{Loss function.}
\Cref{tab:loss_ablation} ablates the consistency objective used for test-time adaptation. Our default objective is the asymmetric KL loss defined in \cref{sec:asymmetric_ttt}, which follows the teacher-student distillation view of matching a student distribution to a fixed teacher distribution~\citep{hinton2015distilling}.
\begin{table}[t]
\centering
\small
\begin{tabular}{lcccccc}
\toprule
Objective & IF base & Image-first TTT & $\Delta_{\mathrm{IF}}$ 
          & QF base & Question-first TTT & $\Delta_{\mathrm{QF}}$ \\
\midrule
Asym.\ KL (ours)  & 61.6 & \textbf{62.3} & \textbf{+0.7} & 35.5 & \textbf{61.5} & \textbf{+26.0} \\
Asym.\ CE         & 61.6 & 62.1 & +0.5 & 35.5 & \textbf{61.5} & \textbf{+26.0} \\
Hard pseudo-label & 61.6 & 61.6 & +0.0 & 35.5 & 60.9 & +25.4 \\
Sym.\ CE          & 61.6 & 61.3 & -0.3 & 35.5 & 45.3 & +9.8 \\
\bottomrule
\vspace{1mm}
\end{tabular}
\caption{
\textbf{Loss-function ablation (InternVL2-8B/MMStar).}
Asymmetric objectives improve question-first accuracy more than symmetric CE. Symmetric CE also slightly reduces image-first accuracy, consistent with the risk of treating the weaker ordering as an equally reliable training signal.
}
\label{tab:loss_ablation}
\end{table}
We compare against three alternatives. Asymmetric CE keeps the same teacher-student direction but replaces KL with cross-entropy:
\[
\mathcal{L}_{\mathrm{asym\text{-}CE}}
=
-\sum_v \operatorname{stopgrad}(p_{\mathrm{IF}}(v)) \log p_{\mathrm{QF}}(v).
\]
Since the teacher distribution is fixed, this has the same gradients as asymmetric KL up to a constant. 

Hard pseudo-labeling also keeps the same direction, but replaces the soft image-first distribution with its argmax prediction~\citep{lee2013pseudo}:
\[
\mathcal{L}_{\mathrm{hard}}
=
-\log p_{\mathrm{QF}}(\hat{y}_{\mathrm{IF}}),
\qquad
\hat{y}_{\mathrm{IF}} = \arg\max_v p_{\mathrm{IF}}(v).
\]
This discards uncertainty in the image-first distribution. 

Symmetric CE follows prior test-time consistency objectives that align predictions across semantically equivalent variants~\citep{chou2026testtime}:
\[
\mathcal{L}_{\mathrm{sym\text{-}CE}}
=
\frac{1}{2}
\left[
-\sum_v p_{\mathrm{IF}}(v)\log p_{\mathrm{QF}}(v)
-
\sum_v p_{\mathrm{QF}}(v)\log p_{\mathrm{IF}}(v)
\right].
\]
Unlike the asymmetric objectives, this treats image-first and question-first as equally reliable signals.

The results support the directional framing in \cref{sec:asymmetric_ttt}: question-first should be adapted toward image-first, rather than forcing both prompt orders toward a shared midpoint.

\paragraph{Layer window.}
\Cref{tab:window_ablation} compares adaptation across different layers. Mid-layers produce the strongest question-first recovery, while early and late windows are less effective. 
\begin{table}[t]
\centering
\small
\begin{tabular}{lcccccc}
\toprule
Window & IF base & Image-first TTT & $\Delta_{\mathrm{IF}}$
       & QF base & Question-first TTT & $\Delta_{\mathrm{QF}}$ \\
\midrule
Layers 5--9   & 61.6 & 61.6 & +0.0 & 35.5 & 60.7 & +25.2 \\
Layers 8--12  & 61.6 & 61.6 & +0.0 & 35.5 & 60.9 & +25.4 \\
Layers 13--17 & 61.6 & 62.1 & +0.5 & 35.5 & 61.0 & +25.5 \\
Layers 16--20 & 61.6 & 61.9 & +0.3 & 35.5 & \textbf{61.7} & \textbf{+26.2} \\
Layers 18--22 & 61.6 & \textbf{62.3} & \textbf{+0.7} & 35.5 & 61.5 & +26.0 \\
Layers 20--24 & 61.6 & 61.9 & +0.3 & 35.5 & 60.7 & +25.2 \\
\bottomrule
\vspace{1mm}
\end{tabular}
\caption{
\textbf{Layer-window ablation on InternVL2-8B/MMStar.}
All windows recover more than 25 points of question-first accuracy. Layers 16--20 gives the largest question-first gain, while layers 18--22 gives the largest image-first gain and is used as the default adaptation window.
}
\label{tab:window_ablation}
\end{table}

\paragraph{Moving versus frozen teacher.} Our default method recomputes the image-first teacher after each TTT update, while stopping gradients through the teacher branch at each step. We compare this to a frozen-teacher variant, where the image-first pseudo-label is computed once from the pretrained model and kept fixed throughout adaptation. \Cref{tab:frozen_vs_moving} shows that recomputing the teacher is critical. With a frozen teacher, question-first accuracy improves from $35.5\%$ to only $43.9\%$. With the moving teacher, question-first accuracy reaches $61.5\%$ at the same step count. The moving teacher also improves image-first accuracy by $+0.7$ points, while the frozen teacher slightly decreases it. This supports the bootstrapping interpretation: as the shared weights improve question-first predictions, the recomputed image-first teacher also becomes sharper, providing a stronger target for later updates.

\begin{table}[t]
\centering
\small
\begin{tabular}{lcccccc}
\toprule
Teacher & IF base & IF TTT & $\Delta_{\mathrm{IF}}$ 
        & QF base & QF TTT & $\Delta_{\mathrm{QF}}$ \\
\midrule
Moving (ours) & 61.6 & \textbf{62.3} & \textbf{+0.7} & 35.5 & \textbf{61.5} & \textbf{+26.0} \\
Frozen        & 61.6 & 61.5 & -0.1          & 35.5 & 43.9          & +8.5  \\
\bottomrule
\end{tabular}
\vspace{0.5cm}
\caption{
\textbf{Moving versus frozen teacher on InternVL2-8B/MMStar.}
Both variants use $K{=}5$ TTT steps and the same mid-layer adaptation window. Recomputing the image-first teacher after each update substantially improves question-first recovery and also preserves image-first accuracy.
}
\label{tab:frozen_vs_moving}
\end{table}
\section{Mechanistic Analysis}
\label{sec:analysis}
 \cref{sec:experiments} shows that order-consistency TTT reduces the question-first gap and slightly improves the image-first baseline too. We now ask where this degradation of question-first appears inside the model, and which internal components change after adaptation. The analysis supports two claims. First, activation patching localizes the ordering failure to a narrow mid-layer band. Second, TTT primarily reduces the image-first/question-first mismatch in late-layer MLP sublayers. We focus on InternVL2-8B in the main text because it exhibits the largest ordering gap in \cref{tab:main_results}; additional Qwen3-VL activation-patching results show the same qualitative localization pattern and are reported in \cref{app:patching_qwen3}, providing evidence that the mechanism generalizes across distinct model families.

\subsection{Activation Patching Localizes the Ordering Failure}
\label{sec:activation_patching}

We use activation patching, following causal mediation and causal tracing methods for localizing transformer computations~\citep{vig2020causal,meng2022locating}, to test where image-first computation can repair question-first computation. For each example, we run the model twice: once with image-first prompting and once with question-first prompting. The image-first run is the donor pass, and the question-first run is the recipient pass. At a chosen transformer layer $L$, we replace the final-token hidden state in the recipient pass with the corresponding hidden state from the donor pass:
\[
h_{\mathrm{QF}}^{(L)} \leftarrow h_{\mathrm{IF}}^{(L)}.
\]
The remaining layers are then run normally, and we read out the final answer.

We report two quantities. First, we measure patched question-first accuracy over all examples. Second, we measure recovery on the fixable subset,
\[
\mathcal{F}
=
\{s : \hat{y}_{\mathrm{IF}}(s)=y(s),\ \hat{y}_{\mathrm{QF}}(s)\neq y(s)\},
\]
where image-first is correct and question-first is wrong. Recovery at layer $L$ is
\[
\mathrm{Recovery}(L)
=
\frac{1}{|\mathcal{F}|}
\sum_{s\in\mathcal{F}}
\mathbf{1}
\!\left[
\hat{y}_{\mathrm{patch}}^{(L)}(s)=y(s)
\right].
\]
This subset isolates examples where the ordering gap produces an error that the image-first hidden state could in principle repair. In our experiments, the fixable subset contains $487$ examples for MMStar, $250$ for RealWorldQA, and $700$ for AI2D.

\begin{figure*}[t]
    \centering
    \begin{subfigure}[t]{0.48\textwidth}
        \centering
        \includegraphics[width=\linewidth]{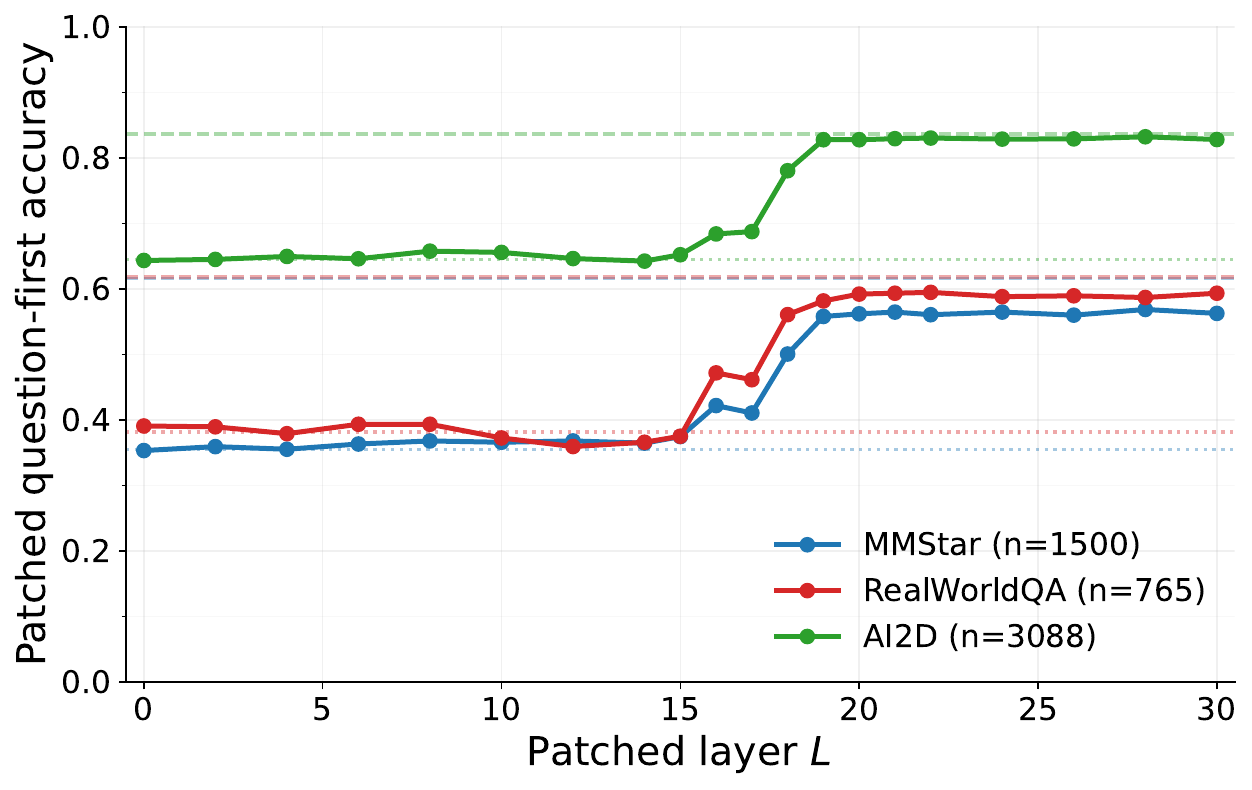}
        \caption{\textbf{Patched accuracy on all examples.}}
        \label{fig:patching_all}
    \end{subfigure}
    \hfill
    \begin{subfigure}[t]{0.48\textwidth}
        \centering
        \includegraphics[width=\linewidth]{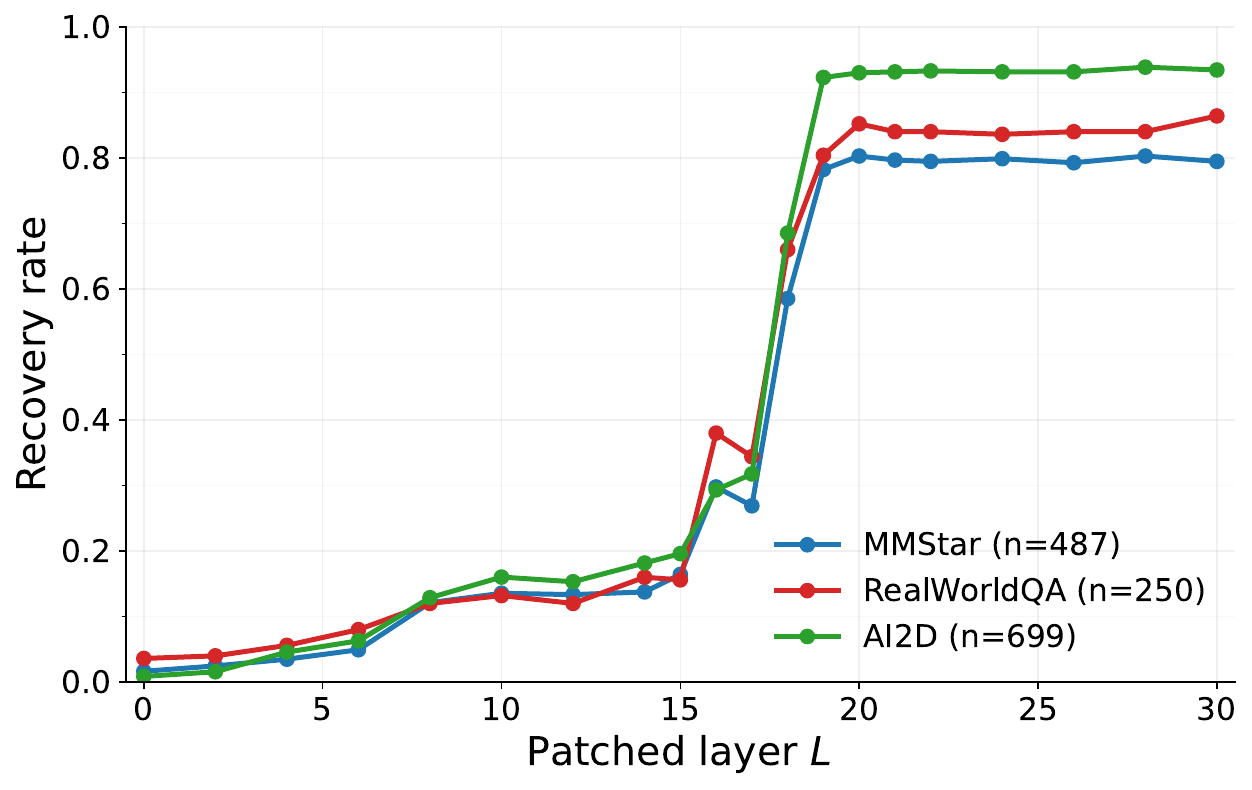}
        \caption{\textbf{Recovery on IF-correct / QF-wrong examples.}}
        \label{fig:patching_fixable}
    \end{subfigure}
    \caption{
    \textbf{Activation patching localizes the ordering failure.}
    We replace the question-first final-token hidden state at layer $L$ with the corresponding image-first hidden state and continue the forward pass normally. Patching becomes effective only in a narrow mid-layer band, where patched question-first accuracy increases and IF-correct/QF-wrong examples are recovered at high rate.
    }
    \label{fig:patching_main}
\end{figure*}

If the two prompt orders diverged uniformly across depth, patching image-first activations into question-first computation would be expected to help across many layers. \Cref{fig:patching_main} shows a different pattern. Panel~(a) reports patched question-first accuracy over all examples, while panel~(b) reports recovery on the fixable subset. Patching early layers has little effect: recovery remains low before layer~17 on all datasets. Around layers~18--20, recovery rises sharply, and after this band the curves plateau. Thus, the ordering failure is not spread uniformly across depth; it is causally concentrated in a narrow mid-layer stage of computation.

\subsection{TTT Aligns Late-Layers MLP Computation}
\label{sec:mlp_alignment}

We next ask whether TTT changes the late-layer components that determine the final prediction. Following \citet{nikankin2025sametask}, we measure overlap between the component sets used by image-first and question-first prompting using normalized intersection-over-union (NIoU). This analysis is restricted to the final layers, so it measures alignment in the decision-stage computation rather than throughout the full network.

For each prompt order, we select the top components used by the forward pass, separately for attention heads and MLP sublayers. Let $S_{\mathrm{IF}}^{c}$ and $S_{\mathrm{QF}}^{c}$ denote the selected component sets for component type $c \in \{\mathrm{heads}, \mathrm{MLP}\}$. We first compute
\[
\mathrm{IoU}^{c}
=
\frac{
|S_{\mathrm{IF}}^{c} \cap S_{\mathrm{QF}}^{c}|
}{
|S_{\mathrm{IF}}^{c} \cup S_{\mathrm{QF}}^{c}|
}.
\]

\begin{figure*}[t]
    \centering
    \includegraphics[width=\linewidth]{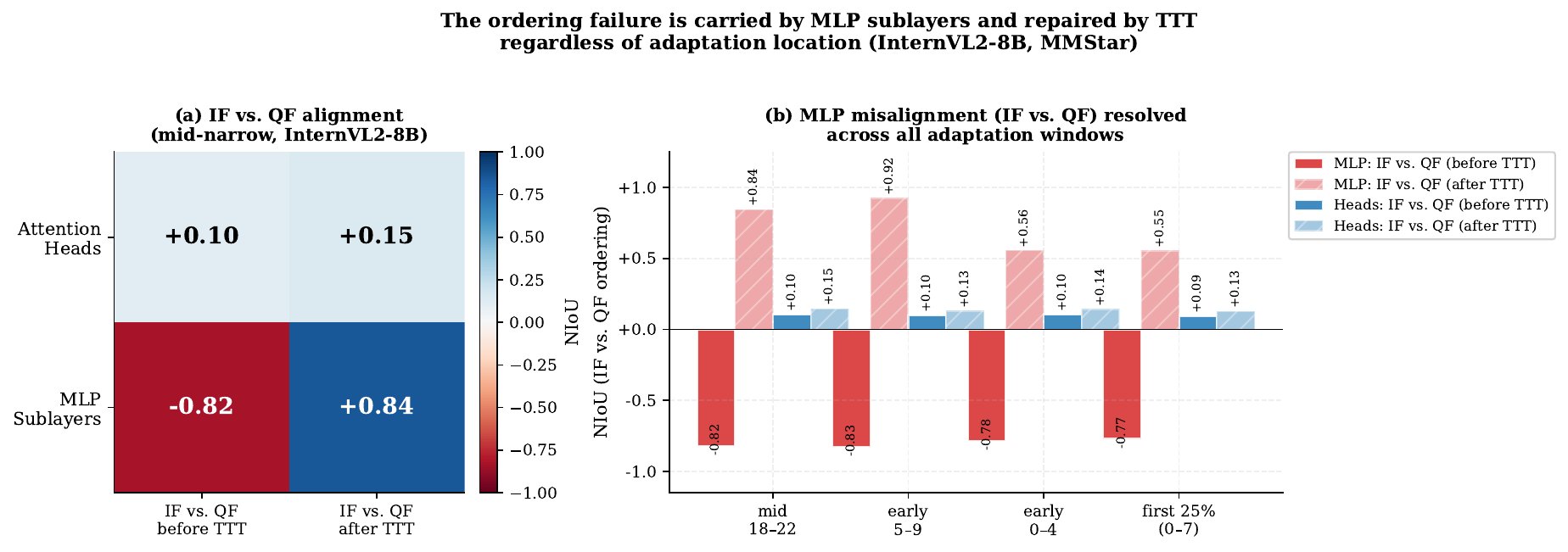}
    \caption{
    \textbf{TTT aligns late-layers MLP computation across prompt orders.}
    We measure normalized intersection-over-union (NIoU) between the top late-layer components used by image-first and question-first forward passes, separately for attention heads and MLP sublayers. Negative NIoU indicates lower overlap than a size-matched random baseline. Before TTT, late-layer MLP sublayers show strongly below-random IF/QF overlap. After TTT, MLP overlap becomes strongly positive across adaptation windows, whereas attention-head overlap changes only modestly.
    }
    \label{fig:mlp_repair}
\end{figure*}

Because raw IoU depends on the number of selected components, we normalize it against a size-matched random baseline:
\[
\mathrm{NIoU}^{c}
=
\frac{
\mathrm{IoU}^{c} - \mathbb{E}[\mathrm{IoU}^{c}_{\mathrm{rand}}]
}{
1 - \mathbb{E}[\mathrm{IoU}^{c}_{\mathrm{rand}}]
}.
\]
Under this normalization, $\mathrm{NIoU}=1$ indicates perfect overlap, $\mathrm{NIoU}=0$ indicates random-level overlap, and $\mathrm{NIoU}<0$ indicates lower overlap than the random baseline.

\Cref{fig:mlp_repair} shows that TTT substantially changes late-layer MLP alignment between the two prompt orders. Before TTT, late-layer MLP components used by image-first and question-first have a strongly below-random overlap, while attention-head overlap is weakly above random. After TTT, the late-layer MLP overlap becomes strongly positive across adaptation windows, whereas attention-head overlap changes only modestly. Thus, TTT also makes the late-layer MLP computation more similar across the two prompt orders, thus improving accuracy.

Together, \cref{fig:patching_main,fig:mlp_repair} provide complementary evidence. Activation patching localizes where the ordering failure can be causally repaired, while late-layer NIoU shows that successful TTT makes the final MLP-mediated computation more order-consistent. This is also consistent with \cref{tab:window_ablation}, where the strongest adaptation windows overlap the layer range where patching becomes effective. Additional drift analyses in Appendix~\ref{app:repair_routes} show that different adaptation windows can reach high output agreement through different internal update footprints.



\section{Discussion, Limitations, and Future Work}
The main implication of our findings is that prompt formatting can induce systematic computational differences, even when the task content is unchanged. This makes modality order a robustness axis for VLM evaluation and a useful source of self-supervision at test time. More broadly, our results suggest that test-time adaptation should not always enforce symmetric consistency: when one input variant is empirically more reliable, the objective should preserve that directionality. The mechanistic evidence further suggests that such failures may be repaired more effectively when adaptation targets the internal region where the inconsistency emerges, rather than updating the model uniformly.
\paragraph{Limitations.}
Our method increases inference cost because each test instance requires two prompt orders and per-instance gradient updates. Our main evaluation focuses on multiple-choice benchmarks, so broader evaluation on captioning, long-form reasoning, and free-form VQA is still needed. Our mechanistic analysis is strongest for InternVL2-8B; additional Qwen3-VL patching results show the same qualitative pattern, but a full cross-family circuit analysis remains future work.
\paragraph{Future work.}
Our ablations and results suggest that image-first gains arise because the moving-teacher update can sharpen shared computation across both orderings. A natural next step is to make this causal account more precise. More broadly, semantically irrelevant ordering changes may expose useful self-supervised signals for improving robustness and accuracy at test time.

\newpage
\bibliography{neurips_2026}
\bibliographystyle{plainnat}

\newpage
\appendix

\section{Text-Only Baseline}
\label{app:textonly}

To verify that both orderings genuinely utilize visual information rather
than language priors, we evaluate InternVL2-8B and Qwen2.5-VL-7B on all
three benchmarks without any image input. Results are shown in
Table~\ref{tab:textonly}. Text-only accuracy is near random chance on
MMStar ($29.2\%$ for InternVL2-8B, $26.6\%$ for Qwen2.5-VL versus
$25\%$ random), confirming strong visual dependency. On AI2D, text-only
accuracy is higher ($59.3\%$ for Qwen2.5-VL), consistent with stronger
language priors in science diagram questions and the smaller ordering gap
observed on AI2D in Table~\ref{tab:main_results}.

\begin{table}[h]
\centering
\small
\begin{tabular}{llccc}
\toprule
Model & Dataset & Text-only & QF baseline & IF baseline \\
\midrule
InternVL2-8B  & MMStar      & 29.2 & 35.5 & 61.6 \\
Qwen2.5-VL-7B & MMStar      & 26.6 & 45.3 & 62.1 \\
Qwen2.5-VL-7B & RealWorldQA & 40.5 & 63.8 & 70.2 \\
Qwen2.5-VL-7B & AI2D        & 59.3 & 69.3 & 84.5 \\
\bottomrule
\end{tabular}
\caption{Text-only accuracy (no image input) compared to question-first
and image-first baselines. On MMStar, text-only accuracy is near random
chance, confirming genuine visual dependency. Higher text-only accuracy
on AI2D reflects stronger language priors in scientific diagram
questions.}
\label{tab:textonly}
\end{table}

\section{CHAIR Hallucination Evaluation}
\label{app:chair}

To test whether ordering-consistency TTT generalizes beyond discriminative
tasks, we evaluate on open-ended image captioning using the CHAIR
metric~\citep{chair}, which measures object hallucination rates in
generated captions. We evaluate InternVL2-8B on $n=500$ COCO images
using a one-sentence captioning prompt. Results are shown in
Table~\ref{tab:chair}.

\begin{table}[h]
\centering
\small
\begin{tabular}{lcc}
\toprule
Condition & CHAIR$_S$ & CHAIR$_I$ \\
\midrule
Image-first (zero-shot)    & 6.8 & 3.9 \\
Question-first (zero-shot) & 7.2 & 4.1 \\
Question-first (TTT)       & \textbf{6.2} & \textbf{3.4} \\
\bottomrule
\end{tabular}
\caption{CHAIR hallucination scores (InternVL2-8B, $n{=}500$ COCO
images, one-sentence prompt, mid-narrow window). After TTT,
question-first CHAIR$_S$ drops from $7.2\%$ to $6.2\%$, falling
below the image-first zero-shot baseline of $6.8\%$. CHAIR$_I$
similarly improves from $4.1\%$ to $3.4\%$.}
\label{tab:chair}
\end{table}

\section{Qwen3 Activation Patching}
\label{app:patching_qwen3}

\begin{figure*}[t]
    \centering

        \centering
        \includegraphics[width=\linewidth]{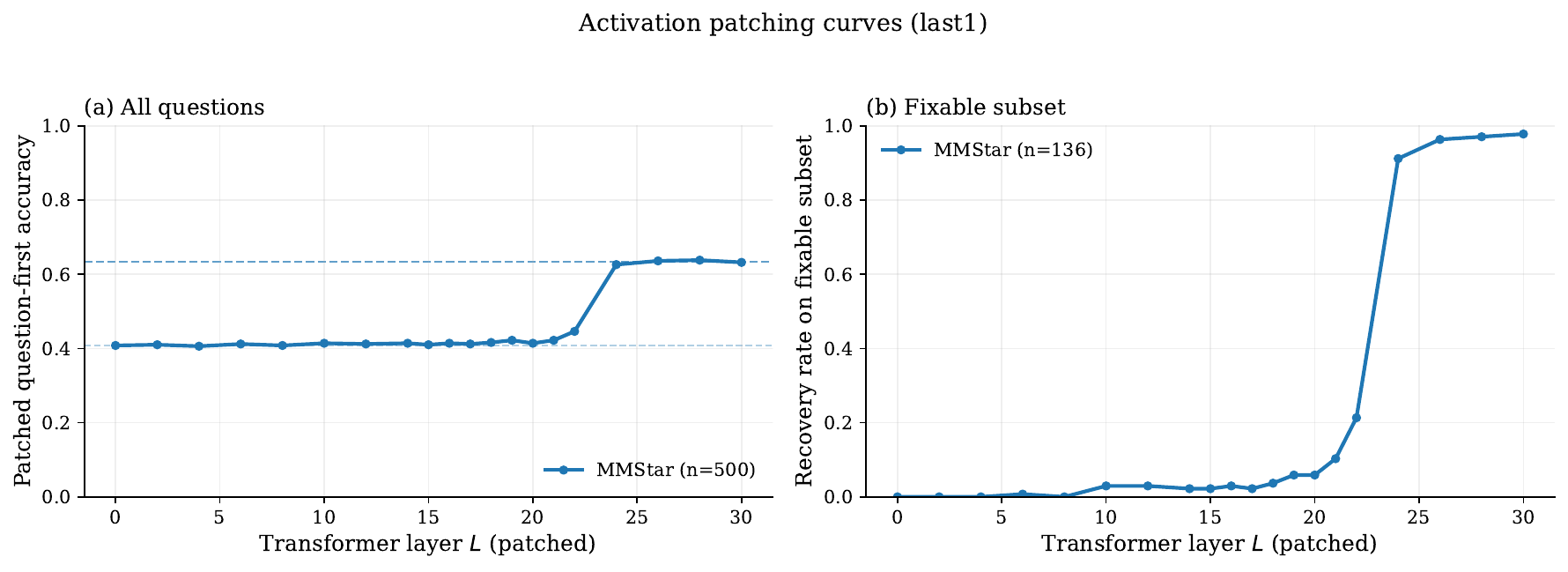}
        \label{fig:patching_all}
    \caption{
    \textbf{Activation patching results - Qwen3.}
    }
    \label{fig:patching_main_qwen}
\end{figure*}

Figure~\ref{fig:patching_main_qwen} shows activation patching results for
Qwen3-VL-8B on MMStar. The same sharp mid-layer localization pattern
observed for InternVL2-8B in Figure~\ref{fig:patching_main} holds for
Qwen3-VL-8B, confirming that the ordering failure is causally localized
to a narrow mid-network band across model families.

\section{Layerwise Repair Routes}
\label{app:repair_routes}
\begin{figure*}[t]
    \centering

    \begin{subfigure}[t]{0.35\textwidth}
        \centering
        \includegraphics[width=\linewidth]{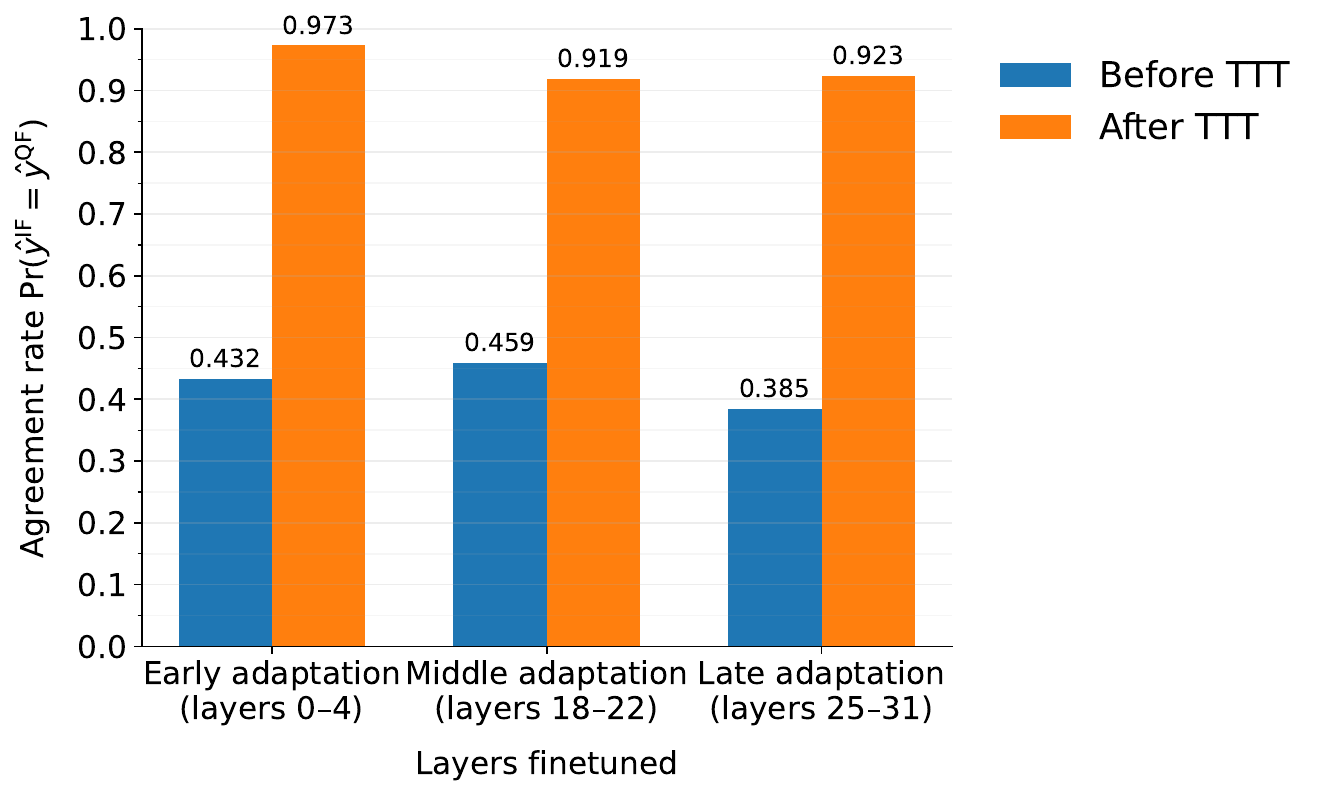}
        \caption{\textbf{Output-level consistency.}}
        \label{fig:agreement}
    \end{subfigure}
    \hfill
    \begin{subfigure}[t]{0.64\textwidth}
        \centering
        \includegraphics[width=\linewidth]{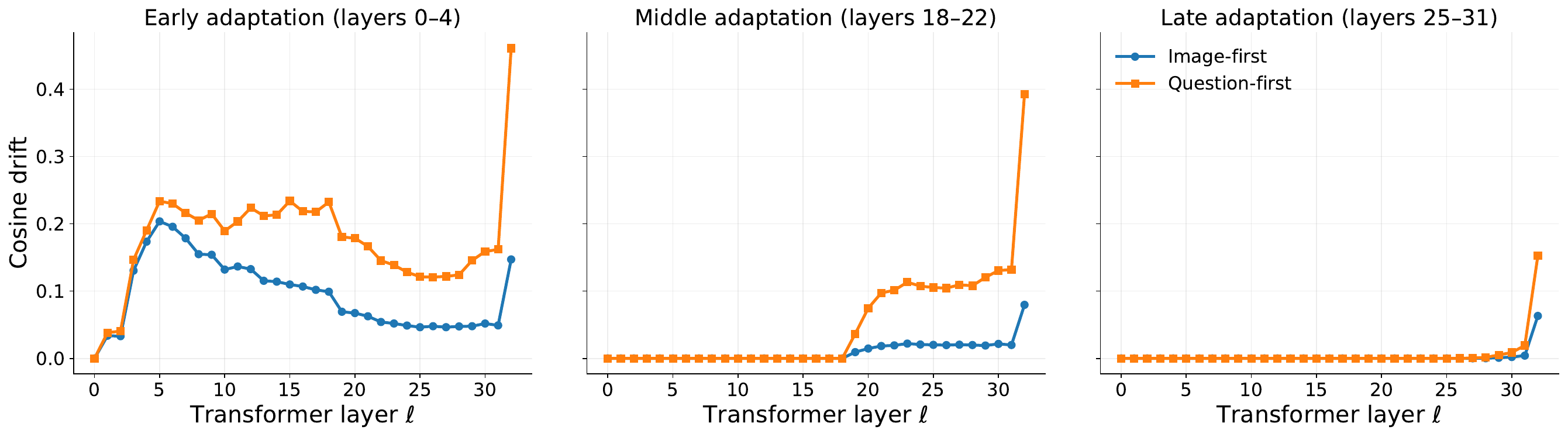}
        \caption{\textbf{Layerwise repair route.}}
        \label{fig:branch_drift}
    \end{subfigure}

\caption{
\textbf{TTT restores order consistency with window-dependent internal drift.}
TTT increases agreement between image-first and question-first predictions for all tested adaptation windows. The corresponding layerwise cosine drift differs by window: early adaptation produces broad changes across the network, middle adaptation concentrates drift near the localized failure band, and late adaptation changes mainly the final layers.
}
\label{fig:repair_routes}
\end{figure*}

Figure~\ref{fig:repair_routes} characterizes how TTT modifies internal
computation under three adaptation windows: early (layers 0--4), middle
(layers 18--22), and late (layers 25--31).

Panel~(a) shows output-level agreement between image-first and
question-first predictions before and after TTT. All three windows
achieve high agreement after adaptation (${\sim}0.92$--$0.97$),
confirming that order consistency can be restored regardless of which
layers are updated.

Panel~(b) shows the layerwise cosine drift --- the change in hidden
states induced by TTT --- separately for image-first and question-first
forward passes. The three windows produce qualitatively distinct internal
footprints. Early adaptation (layers 0--4) produces broad drift
throughout the network for both orderings, indicating that changes at
the input end propagate widely. Middle adaptation (layers 18--22)
concentrates drift near the adapted layers, with question-first drifting
more than image-first in that region --- consistent with a targeted
repair at the causally identified failure site from
Section~\ref{sec:analysis_localization}. Late adaptation (layers 25--31)
produces almost no measurable drift in either ordering, suggesting that
consistency is achieved through minimal weight changes at the final
prediction stage.

These results show that TTT reaches the same output-level goal through
different internal routes depending on where adaptation is applied.
Middle adaptation is the most mechanism-aligned repair: it targets the
layer band identified by activation patching and produces the largest
image-first gain in Table~\ref{tab:window_ablation}.
\label{sec:repair_routes}
\section{Thinking-Mode Ablations}
\label{app:thinking}

\subsection{Effect of Reasoning Mode on Ordering Sensitivity}

We evaluate whether the ordering gap persists when the model is given the
capacity to reason before answering. We compare Qwen3-VL-8B-Instruct
(no reasoning trace) against Qwen3-VL-8B-Thinking, which emits an
explicit chain of thought prior to committing to an answer.

\begin{table}[h]
\centering
\small
\begin{tabular}{llccc}
\toprule
Model & Dataset & IF & QF & Gap (IF$-$QF) \\
\midrule
Instruct (no thinking) & MMStar      & 63.9 & 44.9 & $+19.0$ \\
Thinking               & MMStar      & 69.8 & 62.6 & $+7.2$ \\
\midrule
Instruct (no thinking) & RealWorldQA & 71.6 & 65.5 & $+6.1$ \\
Thinking               & RealWorldQA & 73.8 & 68.8 & $+5.0$ \\
\bottomrule
\end{tabular}
\vspace{0.5cm}
\caption{Ordering gap with and without reasoning mode. Reasoning raises both orderings and does not eliminate the gap. Answers for the thinking model are obtained by generating the full reasoning trace and parsing the committed answer; instruct-model answers are read from
the answer-token distribution.}
\label{tab:thinking-baseline}
\end{table}

The ordering gap is reduced but not closed. This indicates that ordering
sensitivity is not purely a failure of inference-time computation: additional
reasoning recovers part of the deficit in the weaker ordering, yet a residual
structural gap remains.

\subsection{Test-Time Training on the Thinking Model}

We apply the same KL-aligned test-time objective to the
thinking model on MMStar, using $K{=}2$ update steps at layers 18--22 with
$\eta = 5\times10^{-5}$. Because the answer is produced only after the reasoning
trace, all reported accuracies are obtained by generating the full trace and
parsing the committed answer. 
The KL-aligned objective improves the stronger image-first ordering by $+0.040$ while also closing the gap. Accuracies are obtained by
generating the full reasoning trace and parsing the committed answer.

\paragraph{Control.} To confirm that these changes are caused by the test-time
update rather than by variation in decoding, we repeat the procedure with
$K{=}0$ (no gradient step), regenerating both orderings. Accuracy is unchanged:
image-first remains at $72.0$ and question-first at $70.0$, confirming that the
improvement is attributable to the test-time
update and not to decoding variance.

\paragraph{Discussion.} The $K{=}0$ control establishes that test-time updates
causally alter the thinking model's committed answer, despite the answer being
separated from the updated representation by a full reasoning trace. This shows
that the ordering-repair mechanism identified for non-reasoning models extends to
the reasoning-mode setting.




\section{Window Ablation on AI2D}
\label{app:window_ai2d}

Table~\ref{tab:window_ai2d} reports the window ablation for InternVL2-8B
on AI2D. The mid-narrow window (layers 18--22) achieves the best
question-first recovery while preserving image-first accuracy, consistent
with the MMStar results in Section~\ref{sec:experiments}.
\begin{table}[h]
\centering
\small
\begin{tabular}{lcccc}
\toprule
Window & IF base & IF TTT & QF base & QF TTT \\
\midrule
Layers 0--4  (early) & 83.6 & 83.1 & 64.5 & 82.0 \\
Layers 18--22 (mid)  & 83.6 & 83.6 & 64.5 & \textbf{83.0} \\
Layers 25--31 (late) & 83.6 & 83.3 & 64.5 & 80.1 \\
\bottomrule
\end{tabular}
\vspace{0.5cm}
\caption{Window ablation on AI2D (InternVL2-8B). Mid-narrow layers
achieve the strongest QF recovery while preserving IF accuracy,
consistent with MMStar results.}
\label{tab:window_ai2d}

\end{table}

\section{TTT Trajectories}
\label{app:trajectories}

\begin{figure*}[t]
    \centering
    \includegraphics[width=\linewidth]{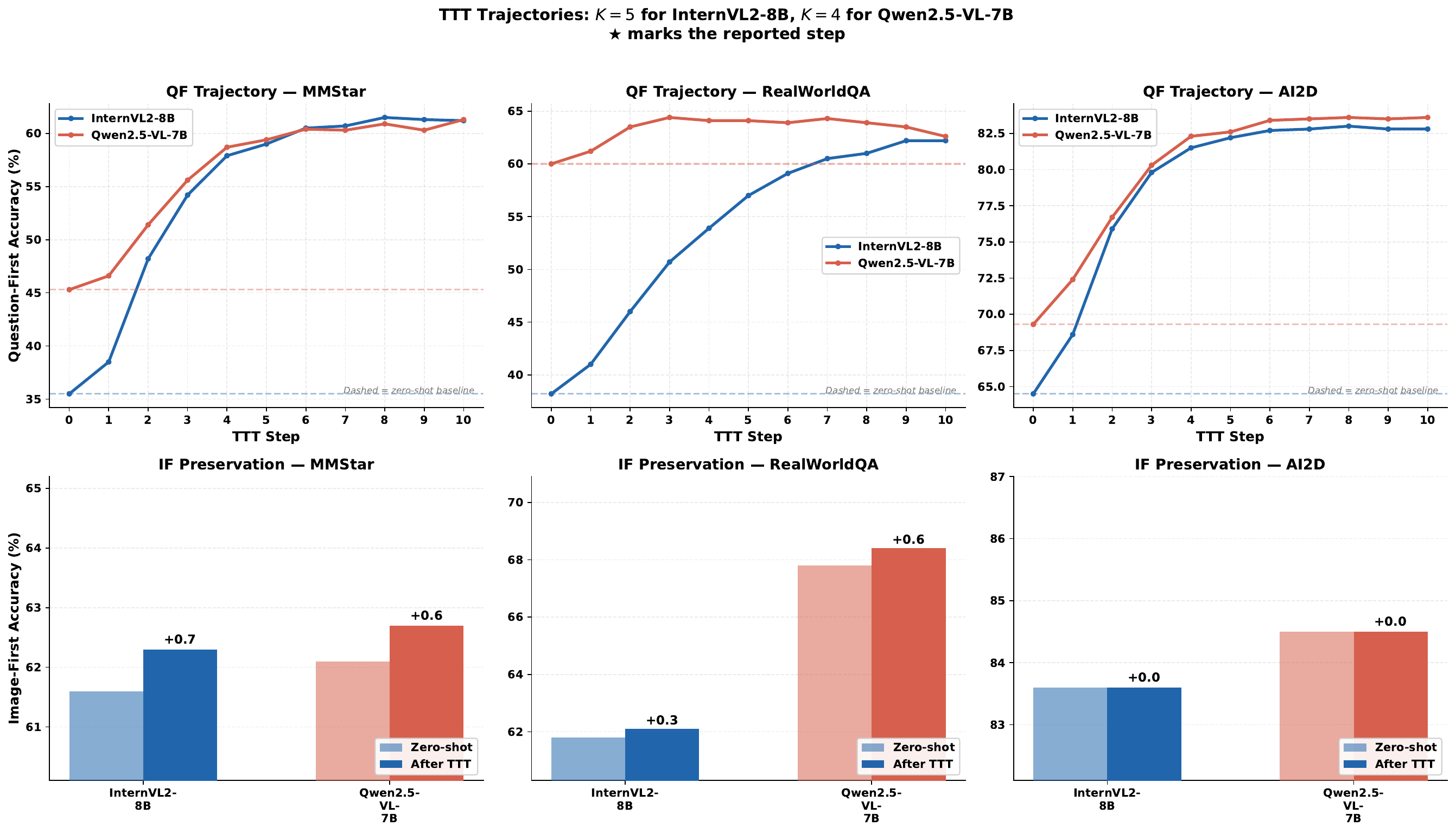}
    \caption{\textbf{TTT trajectories and image-first preservation
    across datasets.} Top row: question-first accuracy across $T{=}10$
    TTT steps for InternVL2-8B ($K{=}5$, blue) and Qwen2.5-VL-7B
    ($K{=}4$, red). Dashed lines indicate zero-shot baselines.
    Bottom row: image-first accuracy before and after TTT at the
    reported step. Image-first accuracy is maintained or slightly
    improved across most settings.}
    \label{fig:appendix_trajectories}
\end{figure*}

Figure~\ref{fig:appendix_trajectories} shows the full TTT trajectories
for InternVL2-8B and Qwen2.5-VL-7B across all three benchmarks. The
top row shows question-first accuracy at each TTT step; the bottom row
shows image-first accuracy at the reported step $K$ compared to the
zero-shot baseline. Question-first accuracy rises consistently across
all settings. Image-first accuracy is maintained or slightly improved
in most settings, with small degradations ($|\Delta_\text{IF}| \leq
0.4$) on AI2D where the ordering gap is smallest.

Our mechanistic analysis focuses on InternVL2-8B; while activation
patching results on Qwen3-VL-8B (Appendix~\ref{app:patching_qwen3})
suggest the localization generalizes, a full mechanistic study across
all three model families remains future work. The window ablation is
evaluated on InternVL2-8B and MMStar only. TTT requires two forward
passes and up to $T=10$ gradient steps per test instance, increasing
inference cost relative to standard prompting; reducing this overhead
is a practical direction for future work. Finally, our evaluation is
restricted to multiple-choice benchmarks and one open-ended captioning
metric; broader evaluation on generative tasks is left to future work.

\section{Broader Impact}
Modality order sensitivity is a systematic and previously underappreciated
failure mode in VLMs that could affect deployed systems where prompt
formatting is not carefully controlled. Our test-time repair requires no
retraining and is applicable to any existing VLM, but increases inference
cost due to two forward passes and per-instance gradient updates. We do
not foresee significant negative societal impacts beyond the general risks
associated with improved VLM capabilities.

\end{document}